\documentclass[conference]{IEEEtran}
\IEEEoverridecommandlockouts
\usepackage{cite}
\usepackage{amsmath,amssymb,amsfonts}
\usepackage{graphicx}
\usepackage{textcomp}
\usepackage{xcolor}
\usepackage{xcolor}         
\usepackage{graphicx}
\usepackage{comment}
\usepackage{multirow,tabularx}
\usepackage{verbatim}
\usepackage{booktabs}
\usepackage{enumitem}
\usepackage{blindtext}
\usepackage{algorithm}
\usepackage{algpseudocode}
\usepackage{amsfonts}
\usepackage{amssymb}
\usepackage{amsmath}
\usepackage{url}

\definecolor{lightpurple}{HTML}{DAD2E9}
\definecolor{lightgreen}{HTML}{D8EAD2}

\def\BibTeX{{\rm B\kern-.05em{\sc i\kern-.025em b}\kern-.08em
    T\kern-.1667em\lower.7ex\hbox{E}\kern-.125emX}}
\begin{document}

\title{AfriBED: a Speech Recognition Bias Evaluation Dataset for Low Resource African Languages\\}

\author{\IEEEauthorblockN{Claytone Sikasote}
\IEEEauthorblockA{\textit{Department of Computer Science} \\
\textit{University of Cape Town}\\
Cape Town, South Africa \\
skscla001@myuct.ac.za}
\and
\IEEEauthorblockN{Hussein Suleman}
\IEEEauthorblockA{\textit{Department of Computer Science} \\
\textit{University of Cape Town}\\
Cape Town, South Africa \\
hussein@cs.uct.ac.za}
\and
\IEEEauthorblockN{Jan Buys}
\IEEEauthorblockA{\textit{Department of Computer Science} \\
\textit{University of Cape Town}\\
Cape Town, South Africa \\
jan.buys@uct.ac.za}
}

\maketitle

\begin{abstract}
Despite advancements in automatic speech recognition systems (ASR), 
current systems exhibit speaker-attribute bias, i.e., their speech recognition performance is biased against certain speakers based on their demographic attributes such as gender, accents, and speaker type.  
However, accurately assessing speaker attribute bias in a given ASR system requires the availability of a bias evaluation dataset with comparable speech across subgroups of a speaker attribute. In this paper, we introduce African speech Bias Evaluation Dataset (AfriBED), a novel speech recognition bias evaluation dataset for two low resource African languages, Bemba and Nyanja, to detect and estimate gender and speaker-type bias for ASR systems. The dataset comprises 9 hours of speech with 3928 utterances recorded by 80 speakers for Bemba, while the Nyanja set includes 6260 utterances with approximately 18 hours of speech recorded by 127 speakers. The results of our baseline experiments demonstrate that the datasets are useful for examining gender and speaker-type attribute bias in the target languages.\footnote{The dataset is publicly available at: \url{https://github.com/csikasote/AfriBED}}
\end{abstract}

\begin{IEEEkeywords}
ASR, Fairness, Bemba, Nyanja, Africa languages
\end{IEEEkeywords}

\section{Introduction}
\label{sec:introduction}
In recent years, there has been a growing interest in developing ASR systems for low resource African languages~\cite{imam2025automatic}. With the majority of languages spoken across the continent classified as low resource languages~(LRLs), most research efforts have focused on building training data dataset in order to improve ASR accuracy~\cite{sikasote-anastasopoulos-2022-bembaspeech,sikasote23_interspeech,sikasote-etal-2023-big,um2025speech}. Although this research address the inherent limitations that characterize LRLs, one research dimension that has not received significant attention is how to effectively evaluate speech recognition models for speaker attribute biases~\cite{sikasote2025sacair}.

In this paper, we present our data collection effort to address the issue of limited speech recognition bias evaluation datasets for African languages~\cite{siminyu-etal-2022-corpus,3769089}. We introduce a novel bias evaluation dataset containing read speech in two low resource African languages: Bemba and Nyanja. The dataset is created for evaluating ASR systems for gender and speaker-type (native and non-native speaker) biases. We carefully constructed the datasets to ensure that the subgroups for each gender and speaker-type attribute have the same speech content to allow fair performance comparison. This is achieved by recording the same read text sentences by speakers of each speaker attribute subtype. To the best of our knowledge, our work is the first to create dedicated ASR bias evaluation datasets to examine speaker-attribute bias in ASR systems for Bemba and Nyanja. 

\section{Related work}
\label{sec:related_work}
Several bias evaluation datasets exist to evaluate ASR systems for various forms of speaker-attribute bias~\cite{kendall_2023_coraal,meyer-etal-2020-artie,veliche24_interspeech, 9747501, jahan25_interspeech}. The most popular is the Artie Bias Corpus~\cite{meyer-etal-2020-artie} based on the Mozilla Common Voice corpus~\cite{ardila-etal-2020-common} for age, gender, and accent bias detection. Liu et al.~\cite{9747501} created the Casual Conversations dataset for gender, age, and skin tone. Shriberg et al.~\cite{shriberg-etal-2004-icsi} curated 72 hours of conversation-like speech based on dialogues from naturally-occuring meetings.  There are others like the recently introduced FairSpeech~\cite{veliche24_interspeech} dataset. However, the majority of these datasets are designed for English ASR systems, leaving the majority of languages without tools to examine speaker attribute bias, particularly LRLs. In addition, the speech across attribute subgroups in these datasets is not the same. The subgroups of attributes do not have speech drawn from the same text sentences, resulting in unfair comparisons. In contrast, our study aims to create bias evaluation datasets for LRLs, targeting African languages. Therefore, we create a novel speech recognition bias evaluation dataset that covers two Sub-Saharan African languages: Bemba and Nyanja, which are spoken predominantly and widely in Zambia and neighboring countries~\cite{zamstats2025census}.

\section{Methodology}
\label{sec:methodology}
In this section, we provide details of the different phases of our data collection process. We highlight that we obtained ethical clearance for data collection and followed all ethical standards for responsible data collection practices.

\subsection{Demographic survey} 
\label{sec:demographic_study}
We started by conducting a demographic survey among students at the University of Zambia, targeting students in the Department of Computing and Informatics using convenient sampling for our selection criteria. The survey was carried out for three reasons: to determine the native languages (amongst languages spoken in Zambia) spoken among the students who are our target population; to estimate the number of students we could invite to participate voluntarily in the successive phases; and to establish the demographic information that we could consider for our bias evaluation dataset. 
The demographic survey was set up using the Google Forms platform and sent out to the students' university email addresses.

A total of 359 students responded to the survey. The responses showed that Bemba (41.23\%) and Nyanja (44.57\%) were the most populous native languages spoken among students and spoken as their first or second language. None of the other languages spoken had a critical mass.
Gender and speaker-type i.e., native and non-native speakers, were the two possible speaker attributes that could be captured for our data collection.~\footnote{We define the first language as the native language the speaker has been exposed to since birth. A native speaker is one where the language is their first language, while a non-native speaker is one where the language is not their first language but they declare a good degree of proficiency in reading, understanding and writing.} We did not find sufficiently large subgroups for other attributes such as race, age and region. All participants identified themselves as black, belonging to a single age group, between 18 and 25 years, and the majority (more than 70\%) came from one region, Lusaka.

\subsection{Generating text sentences}
Due to the lack of high-quality public textual data in our target languages, we obtained 1000 English text sentences from the test split of the open-source ASR dataset, LibriSpeech~\cite{panayotov2015librispeech} and translated them into Bemba and Nyanja for audio recording. The LibriSpeech dataset comprises 1000 hours of speech in the English language derived from publicly domain audio books from the LibriVox project.~\footnote{https://librivox.org/}
To ensure that we obtained high-quality text translations, we recruited two paid professional translators, one translator per language, to translate the English sentences into the target languages. In addition, for each language, we recruited three native speakers (students) on a voluntary basis to verify and ensure that no mismatch occurred in the English-to-translation sentence pairs.  

\subsection{Creating audio recordings}
\begin{figure}[ht!]
\centering
\includegraphics[width=9.00\textwidth, height=4.50cm, keepaspectratio]{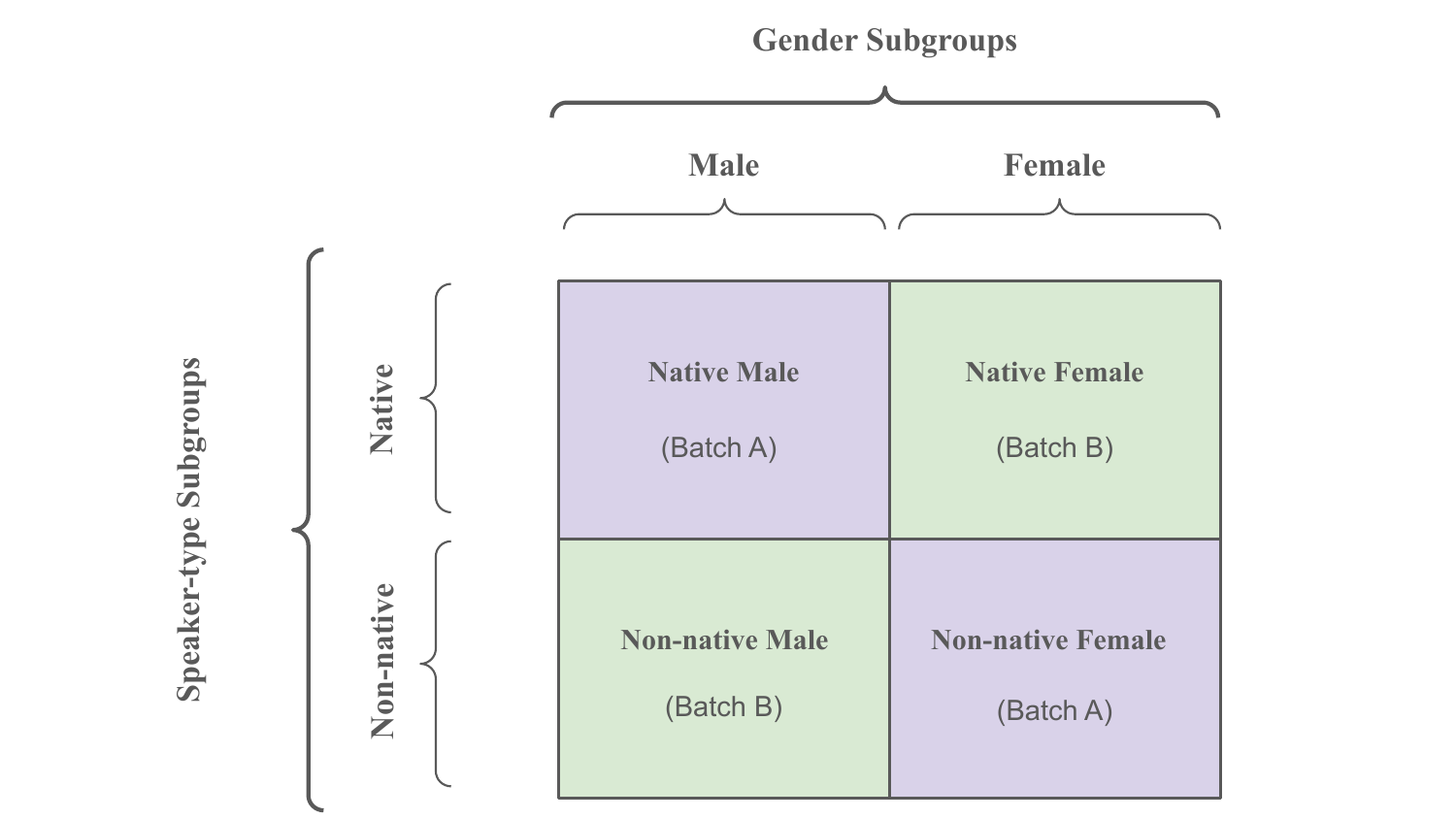}
\caption{Our strategy for allocating text sentences to primary subgroups to create audio recordings, ensuring comparable speech between subgroups of both gender and speaker-type attributes and without speaker overlap.} 
\label{fig:bias_dataset_design}
\end{figure}

 We recruited a total of 205 students to record speech from translated text sentences on a voluntary basis. Most of the students were drawn from the participants in the survey who expressed interest in participating in the audio recording task.  We carefully designed our recording strategy so that the speech between subgroups of the speaker attributes (gender and speaker type) is the same to ensure subgroup comparability. To achieve this, we defined two sets of speaker attribute groups: \emph{primary groups} (\emph{native male}, \emph{non-native male}, \emph{native female}, and \emph{non-native female}) and \emph{secondary groups} (\emph{male}, \emph{female}, \emph{native} and \emph{non-native}). The primary groups are intersectional attribute groups between gender and speaker-type attribute groups, while secondary groups are direct subgroups of gender (male and female) and speaker-type (native and non-native) attributes, whose data composition can be worked out and aggregated from the corresponding primary groups. To assign text sentences for audio recording to participants, we grouped the participants according to their self-identified primary groups. We then split the 1000 text sentences into two batches: \emph{Batch A}, containing sentences 1–500, and \emph{Batch B}, containing sentences 501–1000, and assigned them for recording to each primary group as shown in Fig.~\ref{fig:bias_dataset_design}. Furthermore, we split the text sentences in each batch into portions of 50 text sentences, which we then assigned to each participant to record. To ensure speech variation and diversity, and to allow in-group inter-speaker evaluation, we assigned each 50 sentence block to at least three participants. Using this strategy, gender and speaker-type subgroups can be assigned from the primary groups with speech in the subgroups based on the same sentences and without speaker overlap. The audio files were recorded using the Audacity software, with a sampling rate of 16kHz and 32-bit float PCM sample encoding setting.~\footnote{https://www.audacityteam.org/} All the utterances are within 20 seconds.

\begin{table}[ht!]
\centering
\caption{Summary statistics of the bias evaluation datasets with Native Male (NM), Native female (NF), non-native male (NNM) and non-native female (NNF) primary intersectional speaker attribute subgroups.}
\renewcommand{\arraystretch}{1.0}
\setlength{\tabcolsep}{5pt}
\begin{tabular}{lccc |ccc}
\toprule
\multirow{2}{*}{\textbf{Subgroup}}
& \multicolumn{3}{c}{\textbf{Bemba}} 
& \multicolumn{3}{c}{\textbf{Nyanja}} \\
\cmidrule(lr){2-4} \cmidrule(lr){5-7}
& \textbf{Files} & \textbf{Speakers} & \textbf{Hours}
& \textbf{Files} & \textbf{Speakers} & \textbf{Hours} \\
\midrule
\addlinespace[0.5em]
\multicolumn{2}{l}{\textit{Overall sets}} \\
\midrule
NM         & 1464 & 30 & 3.19 & 2673 & 54 & 7.04 \\
NF       & 821  & 17 & 1.86 & 1144 & 23 & 3.12 \\
NNM     & 1195 & 24 & 3.28 & 1894 & 39 & 6.85 \\
NNF   & 448  & 9  & 0.99 & 549  & 11 & 1.52 \\
\midrule
\textbf{Total} 
& \textbf{3928} & \textbf{80} & \textbf{9.32}
& \textbf{6260} & \textbf{127} & \textbf{18.53} \\
\midrule

\addlinespace[0.5em]
\multicolumn{2}{l}{\textit{Mini-version sets}} \\
\midrule
NM         & 350 & 7  & 0.79 & 299 & 6  & 1.10 \\
NF       & 349 & 7  & 0.73 & 299 & 6  & 0.83 \\
NNM     & 348 & 7  & 0.87 & 300 & 6  & 0.84 \\
NNF   & 349 & 7  & 0.74 & 300 & 6  & 0.80 \\

\midrule
\textbf{Total} 
& \textbf{1396} & \textbf{28} & \textbf{3.13}
& \textbf{1198} & \textbf{24} & \textbf{3.57} \\

\bottomrule
\end{tabular}
\label{tab:all_datasets}
\end{table}

\begin{table*}[t]
\centering
\footnotesize
\setlength{\tabcolsep}{3pt}
\renewcommand{\arraystretch}{1.05}

\caption{Results of fine-tuned ASR models evaluated on overall sets of the language datasets. We report CER on the following subgroups: native male (N), non-native male (NNM), native female (NF), non-native female (NNF), male (M), Female (F), native (N), non-native (NN) and combined (C). Results are averaged over three runs. Lower CER and CER-gap values are better. We report mean and standard deviation (in parentheses) values.}
\label{tab:combined_results}

\resizebox{1.0\textwidth}{!}{%
\begin{tabular}{llccc|ccc|cccc|cccc}
\toprule

\multirow{3}{*}{\textbf{Language}} &
\multirow{3}{*}{\textbf{Model}}
& \multicolumn{3}{c|}{\textbf{Male}}
& \multicolumn{3}{c|}{\textbf{Female}}
& \multicolumn{4}{c|}{\textbf{Gender}}
& \multicolumn{4}{c}{\textbf{Speaker-type}} \\

\cmidrule(lr){3-5}
\cmidrule(lr){6-8}
\cmidrule(lr){9-12}
\cmidrule(lr){13-16}

&& NM & NNM & CER-gap
& NF & NNF & CER-gap
& M & F & C & CER-gap
& N & NN & C & CER-gap \\

\midrule

\multirow{4}{*}{Bemba}
& SSA-HuBERT-L-60K
& 16.76 (0.17) & 19.38 (0.13) & \textbf{2.62 (0.06)}
& 17.43 (0.38) & 21.09 (0.45) & \textbf{3.66 (0.24)}
& 17.95 (0.15) & 18.71 (0.39) & 18.19 (0.19) & 0.77 (0.35)
& 17.00 (0.21) & 19.84 (0.17) & 18.19 (0.18)& \textbf{2.84 (0.10)} \\

& SSA-HuBERT-XL-60K
& 15.45 (0.09) & 18.29 (0.13) & 2.84 (0.05)
& 15.69 (0.34) & 19.85 (0.17) & 4.17 (0.21)
& 16.74 (0.10) & 17.15 (0.27) & 16.87 (0.16)& \textbf{0.41 (0.18)}
& 15.53 (0.17) & 18.70 (0.13) & 16.87 (0.16) & 3.17 (0.05) \\

& MMS-300M
& 20.21 (0.10) & 21.81 (0.02) & \textbf{1.60 (0.09)}
& 21.30 (0.81) & 23.41 (0.20) & \textbf{2.11 (0.63)}
& 20.94 (0.06) & 22.04 (0.59)& 21.29 (0.23) & 1.11 (0.53)
& 20.61 (0.35) & 22.24 (0.07) & 21.29 (0.24) & \textbf{1.63 (0.30)} \\

& MMS-1B-All
& 14.34 (0.18) & 17.55 (0.08) & 3.21 (0.12)
& 14.40 (0.10) & 18.54 (0.06) & 4.14 (0.07)
& 15.80 (0.13) & 15.85 (0.08) & 15.82 (0.11) & \textbf{0.06 (0.08)}
& 14.36 (0.13) & 17.82 (0.08) & 15.82 (0.11) & 3.45 (0.07)\\

\midrule

\multirow{4}{*}{Nyanja}
& SSA-HuBERT-L-60K
& 25.77 (0.19) & 56.09 (1.02) & 30.32 (0.88)
& 25.81 (0.05) & 40.91 (0.88) & 15.10 (0.93)
& 38.58 (0.52) & 30.54 (0.24) & 36.39 (0.43) & \textbf{8.04 (0.34)}
& 25.78 (0.12) & 52.77 (0.98) & 36.39 (0.44) & 26.99 (0.89) \\

& SSA-HuBERT-XL-60K
& 24.89 (0.14) & 54.62 (0.68) & \textbf{29.73 (0.74)}
& 24.72 (0.26) & 39.72 (0.95) & \textbf{14.10 (0.94)}
& 37.46 (0.27) & 29.42 (0.37) & 35.27 (0.30) & \textbf{8.04 (0.13)}
& 24.84 (0.17) & 51.37 (0.74) & 35.27 (0.30) & \textbf{26.53 (0.78)} \\

& MMS-300M
& 29.35 (1.12) & 49.28 (1.52) & \textbf{19.94 (2.56)}
& 28.53 (0.66) & 42.61 (0.60) & \textbf{14.09 (0.64)}
& 37.78 (0.32) & 32.95 (0.57) & 36.46 (0.32) & \textbf{4.83 (0.54)}
& 29.09 (1.00) & 47.83 (1.06) & 36.46 (0.32) & \textbf{18.74 (1.95)} \\

& MMS-1B-All
& 22.35 (0.08) & 55.63 (0.24) & 33.27 (0.20)
& 21.19 (0.09) & 38.80 (0.29) & 17.60 (0.21)
& 36.41 (0.13) & 26.72 (0.15) & 33.76 (0.14) & 9.69 (0.04)
& 21.99 (0.08) & 51.95 (0.25) & 33.76 (0.14) & 29.96 (0.19) \\

\bottomrule
\end{tabular}%
}
\end{table*}

\subsection{Data validation}
\label{sec:data_validation}
We validated all data collected following the strategy of Conneau et al.~\cite{conneau2023_fleurs} to ensure that only clean and high quality audio recordings were retained. All valid recordings, from either quiet or noisy environments, are kept as is.

\subsection{Resulting Datasets}
\label{sec:resulting_datasets}
Table~\ref{tab:all_datasets} provides summary counts of audio files, number of speakers, and size (in hours) for each subgroup in the resulting evaluation sets. The overall datasets comprise a total of 9.32 hours of Bemba speech (3,928 audio files from 80 speakers) and 18.53 hours of Nyanja speech (6,260 audio files from 127 speakers). In addition, we curated subsets of the overall datasets into mini-version datasets, which include only speech recordings for text segments with at least one contribution from each subgroup, ensuring balanced and comparable evaluation across subgroups. The mini-version datasets include 3.13 hours of Bemba speech (1,396 files from 28 speakers) and 3.57 hours of Nyanja speech (1,198 files from 24 speakers). The dataset, along with transcriptions and anonymized metadata files, are released publicly to the research community under a Creative Commons BY-NC license.\footnote{The dataset is publicly available at: \url{https://github.com/csikasote/AfriBED}}

\section{Baseline Experiments}
\label{sec:experiments_and_results}
To provide baseline speech recogntion and fairness performances of speech models in the target languages, we fine-tune two popular open-source pre-trained speech models: massively multilingual model for speech (MMS)~\cite{pratap2023scalingspeechtechnology1000} and the Africa-centric self-supervised multilingual speech model for Sub-Sahara African languages (SSA-HuBERT-Large-60K)~\cite{caubriere2024africacentric} on existing open-source labeled ASR datasets for Bemba (28 hours) and Nyanja (25 hours) from Zambezi Voice~\cite{sikasote23_interspeech}. The datasets are not balanced with speaker-type and gender attribute speech. We use the 300M and 1B parameter variants of both models. 
We report character error rate (CER) metric~\cite{kurimo06_interspeech} for ASR performance. For bias, we use the absolute difference in ASR performance between the subgroups of the respective speaker attributes (CER-gap)~\cite{FENG2024101567}. 
We use our full collected dataset to evaluate the fine-tuned speech models. 

\section{Results}
\label{sec:results}
Table~\ref{tab:combined_results} provides the results of our baseline experiments. Below we provide key observations, and also compare to the findings of Sikasote et al.~\cite{sikasote2025sacair}, who evaluated gender bias for Bemba and Nyanja ASR on test sets without overlapping sentences between the male and female speech.  

\subsection{Bias Analysis for Bemba}
\label{sec:bemba_results}
 We observe performance disparities across all speaker attribute categories for all fine-tuned models. All models perform better in recognizing speech for native than non-native speakers, suggesting potential bias against non-native speakers in the models. All models perform better in recognizing male than female speech with relatively low CER-gap compared to other attributes. In contrast, Sikasote et al.~\cite{sikasote2025sacair} found inconsistent bias patterns between two evaluation sets both without sentence overlap between the subgroups.

\subsection{Bias Analysis for Nyanja}
\label{sec:nyanja_results}
 We observe significantly higher CER and CER-gap values in Nyanja than in Bemba. We attribute this to the possible impact of background noise. Our speech analysis revealed that approximately 1.5 hours of the total Nyanja data is non-speech compared to Bemba, with only 45 minutes. Similarly to Bemba for the \textit{Speaker Type} attribute, all models perform better in recognizing speech for native speakers than non-native speakers, suggesting potential bias against non-native speakers in the models. For gender, all fine-tuned models exhibit bias against male speakers, consistent with the pattern observed in the study by Sikasote et al.\cite{sikasote2025sacair}.

\section{Conclusion}
\label{sec:conclusion}
This paper introduces AfriBED, a bias evaluation dataset created to examine gender and speaker-type bias in ASR systems for two low resource African languages: Bemba and Nyanja. The dataset's key innovation is that the subgroups of each speaker attribute have the same speech content, making the subgroups comparable for bias evaluation. The results of our baseline experiments demonstrate the usefulness of our datasets to examine gender and speaker-type attribute bias in ASR models for Bemba and Nyanja. In the future, we plan to expand our dataset to more languages, and to add additional speaker attributes. 

\section*{Acknowledgment}
This project would not have been successful without the voluntary participation of students and the incredible support of academic staff in the Department of Computing and Informatics at the University of Zambia. We express our profound gratitude to Mofya Phiri, Monica M. Kabemba, and Alinani Simukanga for the useful conversations and support during the course of the data collection process.

This work was financially supported by the Hasso Plattner Institute (HPI) for Digital Engineering, through the HPI Research School at the University of Cape Town. This work is based on research supported in part by the National Research Foundation of South Africa (Grant Number 151601) and Telkom Centre of Excellence. Computations were performed using facilities provided by the University of Cape Town’s ICTS High Performance Computing (HPC) team:~\url{https://ucthpc.uct.ac.za/}

\section*{AI-Generated Content Disclosure} We used ChatGPT to generate LaTeX code to format and draw tables. The writing, technical content, experiments, and conclusions were produced and verified by the authors.

 \bibliographystyle{ieeetr}
 \bibliography{mybibliography}

\end{document}